\documentclass[conference]{IEEEtran}
\pdfoutput=1
\usepackage{cite}
\usepackage{amsmath,amssymb,amsfonts}
\usepackage{algorithmic}
\usepackage{graphicx}
\usepackage{textcomp}
\usepackage{xcolor}
\usepackage{multirow}
\usepackage{adjustbox}
\usepackage[english]{babel}
\def\BibTeX{{\rm B\kern-.05em{\sc i\kern-.025em b}\kern-.08em
    T\kern-.1667em\lower.7ex\hbox{E}\kern-.125emX}}

\newcommand{\mix}{\ensuremath{\text{Mix}}}

\begin{document}

\title{
Establishing baselines and introducing TernaryMixOE for fine-grained out-of-distribution detection
}


\author{\IEEEauthorblockN{1\textsuperscript{st} Noah Fleischmann}
\IEEEauthorblockA{\textit{Khoury College of Computer Science} \\
\textit{Northeastern University}\\
Boston, MA, United States \\
fleischmann.n@northeastern.edu}
\and
\IEEEauthorblockN{2\textsuperscript{nd} Walter Bennette}
\IEEEauthorblockA{\textit{Air Force Research Lab} \\
Rome, NY, United States \\
walter.bennette.1@us.af.mil}
\and
\IEEEauthorblockN{3\textsuperscript{rd} Nathan Inkawhich}
\IEEEauthorblockA{\textit{Air Force Research Lab} \\
Rome, NY, United States \\
nathan.inkawhich@us.af.mil}
}

\maketitle

\begin{abstract}
Machine learning models deployed in the open world may encounter observations that they were not trained to recognize, and they risk misclassifying such observations with high confidence. Therefore, it is essential that these models are able to ascertain what is in-distribution (ID) and out-of-distribution (OOD), to avoid this misclassification. In recent years, huge strides have been made in creating models that are robust to this distinction. As a result, the current state-of-the-art has reached near perfect performance on relatively coarse-grained OOD detection tasks, such as distinguishing horses from trucks, while struggling with finer-grained classification, like differentiating models of commercial aircraft. In this paper, we describe a new theoretical framework for understanding fine- and coarse-grained OOD detection, we re-conceptualize fine-grained classification into a three part problem, and we propose a new baseline task for OOD models on two fine-grained hierarchical data sets, two new evaluation methods to differentiate fine- and coarse-grained OOD performance, along with a new loss function for models in this task.
\end{abstract}


\section{Introduction}
\label{sec:intro} 

When deployed in the real world, machine learning systems are prone to giving high-confidence predictions when fed inputs outside of their training distribution \cite{Amodei_Olah_Steinhardt_Christiano_Schulman_Mane_2016}. Because of this, an enormous amount of energy has been dedicated to creating models and detection systems that are robust to this kind of error, including many `outlier exposure' methods that include out-of-distribution (OOD) images at training time \cite{OE, ODIN, Energy}. While progress has been significant, many of these models are evaluated on benchmarks that are easily separable because of the strong distinction between the in-distribution (ID) classes and the contrast between the ID and OOD test sets.  For example common benchmarks include the CIFAR10 and CIFAR100 \cite{Cifar10} data sets for in-distribution and coarse-grained data sets like Street View House Numbers (SVHN) \cite{svhn} or Places365 \cite{places} as out-of-distribution. This has led to near perfect performance and increasingly small gains for new methodologies \cite{100percentNormalizingFlow}.

\begin{figure}
\begin{center}
\begin{tabular}{c}
\includegraphics[width=\linewidth]{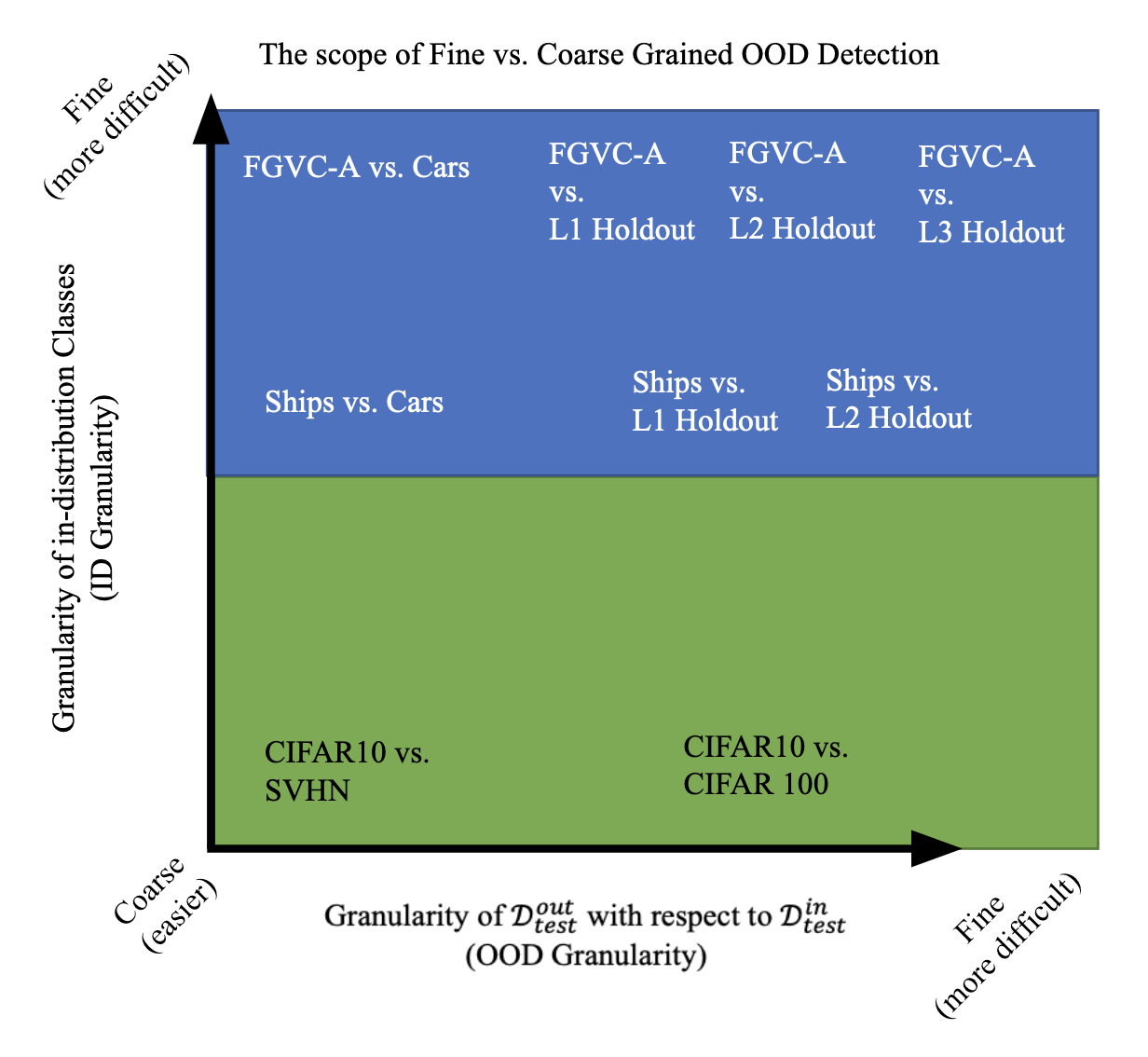}
\end{tabular}
\end{center}
\label{fig:datasetgraph}
\caption{Our understanding of Fine-grained OOD detection. On the horizontal axis we have OOD granularity, the difficulty of detecting samples from the OOD test set relative to the ID set. For example, detecting Street View House Numbers (SVHN) examples when CIFAR10 is the in-distribution set is relatively easier than when CIFAR100 is the out-of-distribution set. Similarly, detection of the three holdout sets described in \ref{sec:fgvc} exhibit a gradient of difficulty in this metric. On the vertical axis, we identify the in-distribution granularity, or the difficulty of the `core task`. For example, \textit{Airplane} is a single class of CIFAR10 while the in-distribution task of FGVC-A is to distinguish 102 types of airplane variants. Existing research has largely focused on OOD detection in the lower half of this graph, shown in green. With this paper, we hope to redefine the problem space in the upper half, shown in blue.}
\end{figure} 

At present, there is little theoretical basis to measure or describe the granularity of the OOD detection task. While it can be measured empirically by training the same architecture on the same objective, and comparing performance on different OOD detection data sets, this method is lacking in a number of respects: it is computationally expensive, dependant on hyperparameter performance, and can only be done \textit{post hoc}.

Our framework for describing data set granularity shown in Fig. \ref{fig:datasetgraph} divides the concept into two parts, which we will call OOD granularity and ID Granularity, that operate independently of one another. OOD granularity, shown on the horizontal axis, refers to the level of precision required to separate the OOD test set, $D^{out}_{test}$, from the ID test set, $D^{in}_{test}$. This increases when the in- and out-of-distribution sets become more visually similar, requiring the detection of similar objects from similar points of view. For example,  using CIFAR10 as an in-distribution set with SVHN and CIFAR100 as out-of-distribution test sets. SVHN is of close cropped photos of house number plates taken from Google Street View making it a coarse-grained OOD task in comparison to CIFAR100, containing finer-grained labels of objects in classes similar to CIFAR10 such as vehicles and aquatic mammals. In the extreme, OOD granularity can be boosted when $D^{out}_{test}$ is composed of a set of hold-out classes of $D^{in}$, as we will outline in Sec. \ref{sec:datasets}.

Similarly, ID granularity, shown on the vertical axis, refers to the precision required to separate the in-distribution classes, describing the difficulty of the core task. This can increase if the classes of one data set compose to a single class in another data set. For example, the first fine-grained dataset we will propose, Fine Grained Visual Classification of Aircraft (FGVC-A) \cite{fgvc}, requires more precision when classifying examples compared to CIFAR10, as the former is identifying airplanes of different types while the later groups all airplanes into a single class. ID granularity is harder to compare if two data sets use disjoint classes, and so comparisons must rely on qualitative human perceptions, or the empirical methods outlined above. 

Crucially, existing research has largely focused on coarse-grained data sets in both of these metrics, shown in the green region on Fig. \ref{fig:datasetgraph}. 

\vspace{5pt}
\noindent In this work we make the following contributions:

\begin{itemize}
    \item \textit{Establish a new baseline for a more difficult OOD detection scenario:} To supersede existing CIFAR benchmarks, we propose training on two fine-grained hierarchical classification data sets, using larger and more detailed images and finer grained in-distribution classes.
    \item \textit{Propose new metrics to evaluate this task:} In order to better measure the performance of fine-grained OOD models, we introduce Hierarchical AUROC/FPR and Semantic vs. True AUROC/FPR .
    \item \textit{Compare five existing methods, and introduce a logical extension of MixOE.}
\end{itemize}

\section{Data sets}
\label{sec:datasets}

\begin{figure}
\begin{center}
\begin{adjustbox}{width=\linewidth}
\begin{tabular}{c}
\includegraphics[width=\linewidth]{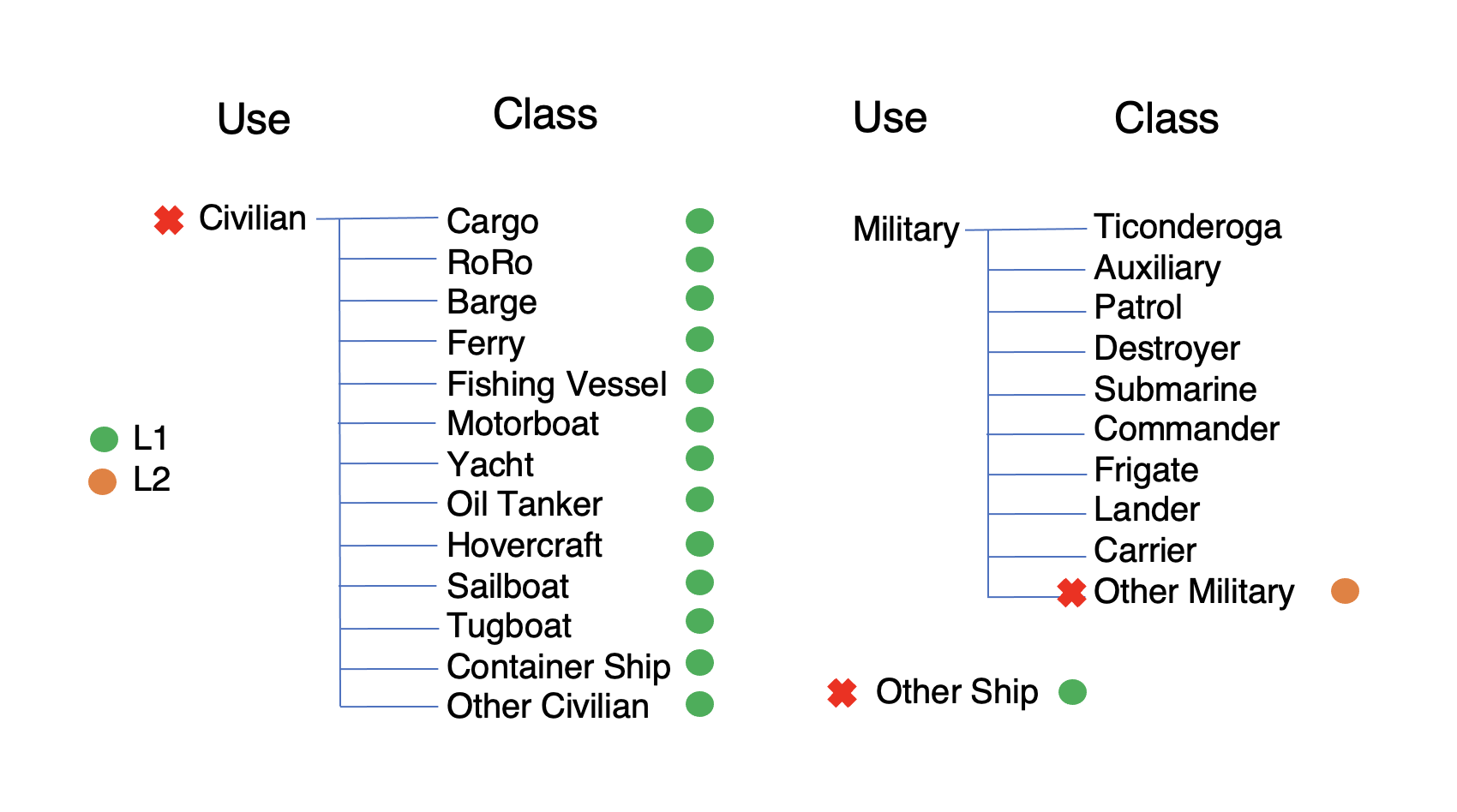}
\end{tabular}
\end{adjustbox}
\end{center}
\caption 
{Our proposed ID/OOD split for ShipsRSImageNet Level 1 OOD classes, held out at the use level, are identified in green. Level 2 classes, held out at the class level, are shown in orange. There is no Level 3 OOD.}
\label{fig:shipssplit}

\end{figure} 

In this section we introduce two fine-grained data sets to establish new baseline classification and OOD detection tasks. Each dataset is constructed with a hierarchical label space, in which certain classes are `held out' to be used as OOD samples during test time. Classes are identified according to the level of the hierarchy at which they are held out such that all downstream classes are OOD. For example, Fig. \ref{fig:shipssplit} shows our proposed split for ShipsRSImageNet \cite{ShipsRS}, where all civilian ship classes are held out as Level 1 OOD. All classes under `civilian' in the hierarchy are in a distinct OOD class (coarser) from the Level 2 `other military' (finer) examples. This allows us to train flat models with no internal hierarchy conceptualization on the finest-grained classes but still evaluate on a gradient of OOD granularity in our Hierarchical AUROC metric outlined in Sec. \ref{sec:hauroc}

\subsection{Fine-grained Visual Classification (FGVC) of Aircraft }
\label{sec:fgvc}

\begin{figure*}[ht]
\begin{center}
\begin{tabular}{c}
\includegraphics[width=0.8\textwidth]{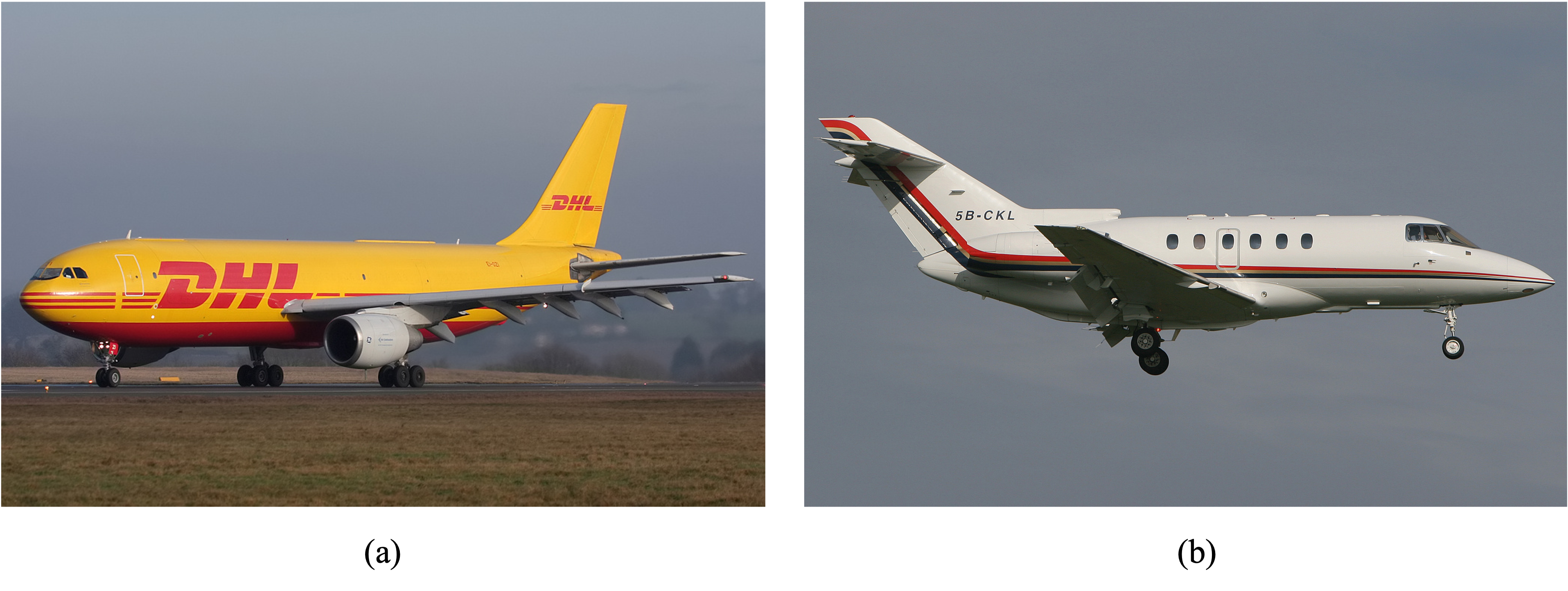}
\end{tabular}
\end{center}
\caption 
{ \label{fig:airplanesample}
Two images from FGVC-Aircraft, with the 20 pixel identifier cropped, (a) a Boeing 747-400 and (b) an Airbus A330-200} 
\end{figure*}

FGVC-Aircraft is a fine-grained hierarchical classification data set composed of 10,200 images of aircraft taken from a ground level point-of-view. While the classification hierarchy contains four levels, only three are visually distinct enough to present a plausible OOD detection task. From coarsest to finest they are: Manufacturer,\textit{ e.g. Boeing}; Family \textit{e.g. Boeing 747}; and Variant \textit{e.g. Boeing 747-200}. There are 40 manufacturers, 70 total families, and 102 total variants. The images vary in size and aspect ratio but all have a resolution of approximately 1-2 megapixels \cite{fgvc}.

\begin{figure*}
\begin{center}
\begin{tabular}{c}
\includegraphics[height=10cm]{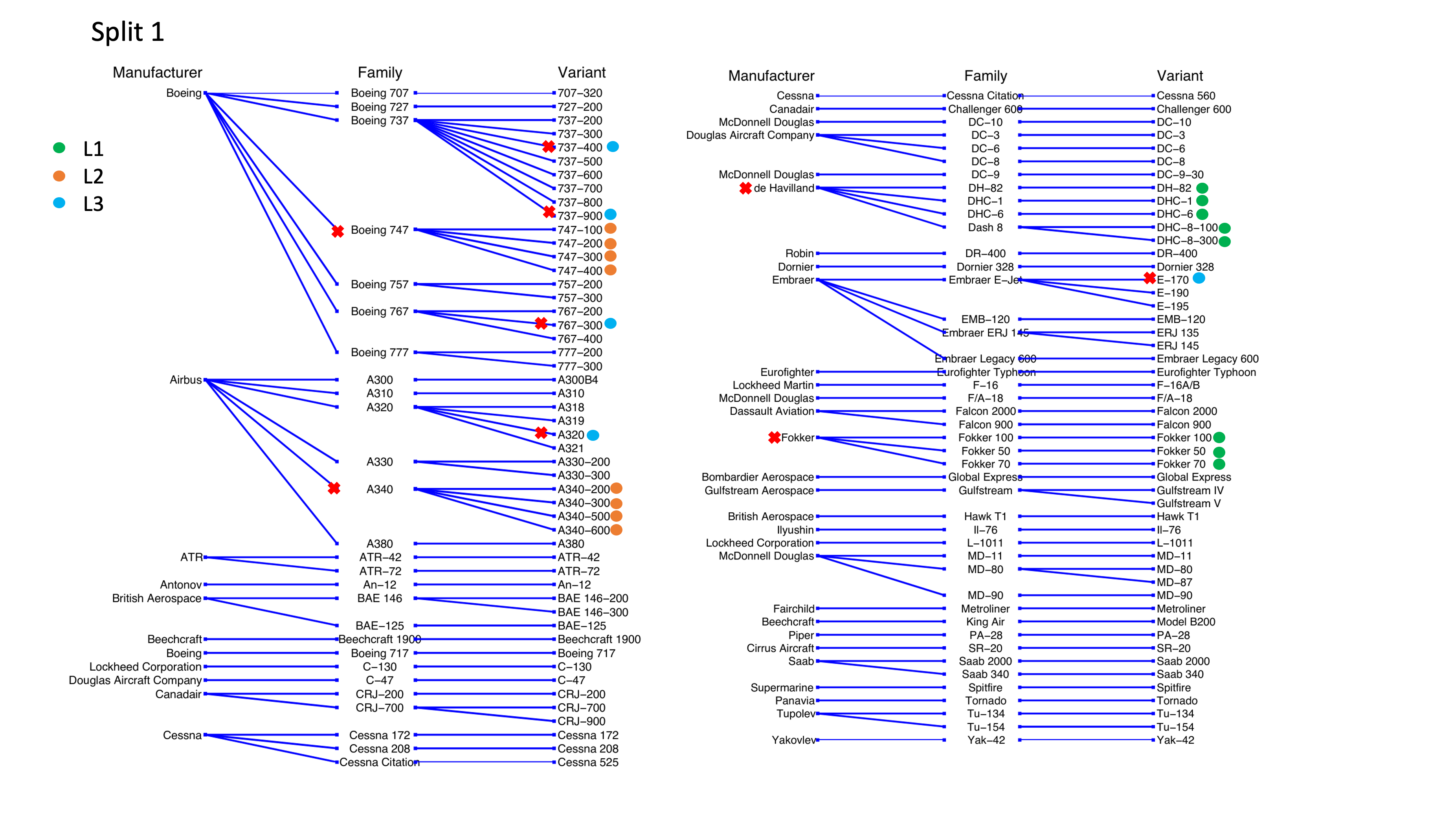}
\end{tabular}
\end{center}
\caption 
{Split 1, taken from \cite{mixup}, shown on a figure from \cite{fgvc}. Level 1 OOD classes, held out at the manufacturer level, are identified in green. Level 2 classes, held out at the model level, are shown in orange. Level 3 classes, held out at the variant level, are shown in blue.}
\label{fig:aircraftsplit}
\end{figure*}

FGVC-Aircraft is split four ways according to \cite{mixup}, each with a hold out test set. As mentioned above, classes are held out at all three levels of the hierarchy with all downstream classes labeled as OOD accordingly. For example, split one of \cite{mixup}, shown in Fig. \ref{fig:aircraftsplit} identifies Boeing 737-400 as a Level 3 OOD hold out class, all Airbus A340 variants (A340-200, A340-300, etc.) as Level 2 OOD, and all de Havilland models (DH-82, DHC-1, etc.) as Level 1 OOD.

\subsection{ShipsRSImageNet}

ShipsRSImageNet is a hierarchical remote-sensing object-detection data set of 13,065 ships in 3,435 images taken from a variety of satellites and sensors. To adapt the data set to a classification task, each ship is cropped to a small area around the given bounding box. To limit the task to just the classification of ships, the broadest level of the hierarchy, which includes the docks surrounding the ships, is excluded. The remaining hierarchy is composed of three levels, of which two are useful for creating a fine-grained OOD task. The finest grained level is excluded due to the heavy imbalance in the number of samples per class (within the Military category, they range from 18 samples for the Midway aircraft carrier to 526 for the Arleigh Burke class destroyers). The two remaining levels are `use' (military versus civilian) and `class' (e.g. Submarine, Aircraft Carrier; Sailboat, Fishing Vessel) \cite{ShipsRS}. The hierarchy for ShipsRSImageNet is not as rich as FGVC-Aircraft, but it provides an opportunity to evaluate methods at an intermediate level of granularity compared to existing CIFAR baselines.

Unlike FGVC-Aircraft, there is no academic consensus or existing literature on how to split the classes of ShipsRSImageNet for fine-grained OOD detection. To alleviate this, we propose a split, shown in Fig. \ref{fig:shipssplit}, between the two classes under use, holding all civilian ships and `other ships' as Level 1 OOD, the `other military' catch-all class as Level 2 OOD and keeping all identified military classes as ID.

\section{Methodology}
\label{sec:method}

\subsection{Evaluation Metrics}
\label{sec:evalmetrics}

\subsubsection{Hierarchical AUROC/FPR}
\label{sec:hauroc}

As we described in Sec. \ref{sec:intro}, there is little theoretical basis for identifying one OOD set as more difficult than another without first training a model and measuring the performance empirically. To address this issue, we can take advantage of the hierarchical nature of our proposed data sets. Our hold out set, $D_{test}^{out}$, has samples held out at each level of the hierarchies in sufficient quantities to be test sets of their own. Using this, we can divide $D_{test}^{out}$ according to hold out level - creating distinct non-overlapping test sets for Level 1, Level 2, and Level 3 – and evaluate separately, thereby creating a suite of metrics that showcase the performance of the model on a gradient of fine-grained classification difficulty. Later, we will confirm this intuition by showing empirically that OOD detection on examples in the Level 3 holdout set is more difficult than detection of Level 2 or Level 1 examples. 

Specifically, we will show that the area under the Receiver Operating Characteristic (AUROC) \cite{AUROC} and the false-positive rate at a 95\% true-positive rate (FPR) will decrease and increase respectively as the hold out set moves deeper into the hierarchy and OOD detection becomes more difficult. Additionally, evaluating on the gradient provides more information than a single hold-out set at the finest level. For example, it is possible for different methods to degrade performance at different rates. By only measuring the AUROC/FPR on the entire hold out set or only on the Level 3 fine-grained set, we miss the possibility that some methods perform better at finer-grained levels despite performing worse overall. Such a model would be preferable in applications where fine-grained OOD detection is most important and users are willing to sacrifice overall performance for this improvement.

\subsubsection{Semantic vs. True AUROC/FPR}
\label{sec:stauroc}

\begin{figure}
\begin{center}
\begin{tabular}{c}
\includegraphics[width=1\linewidth]{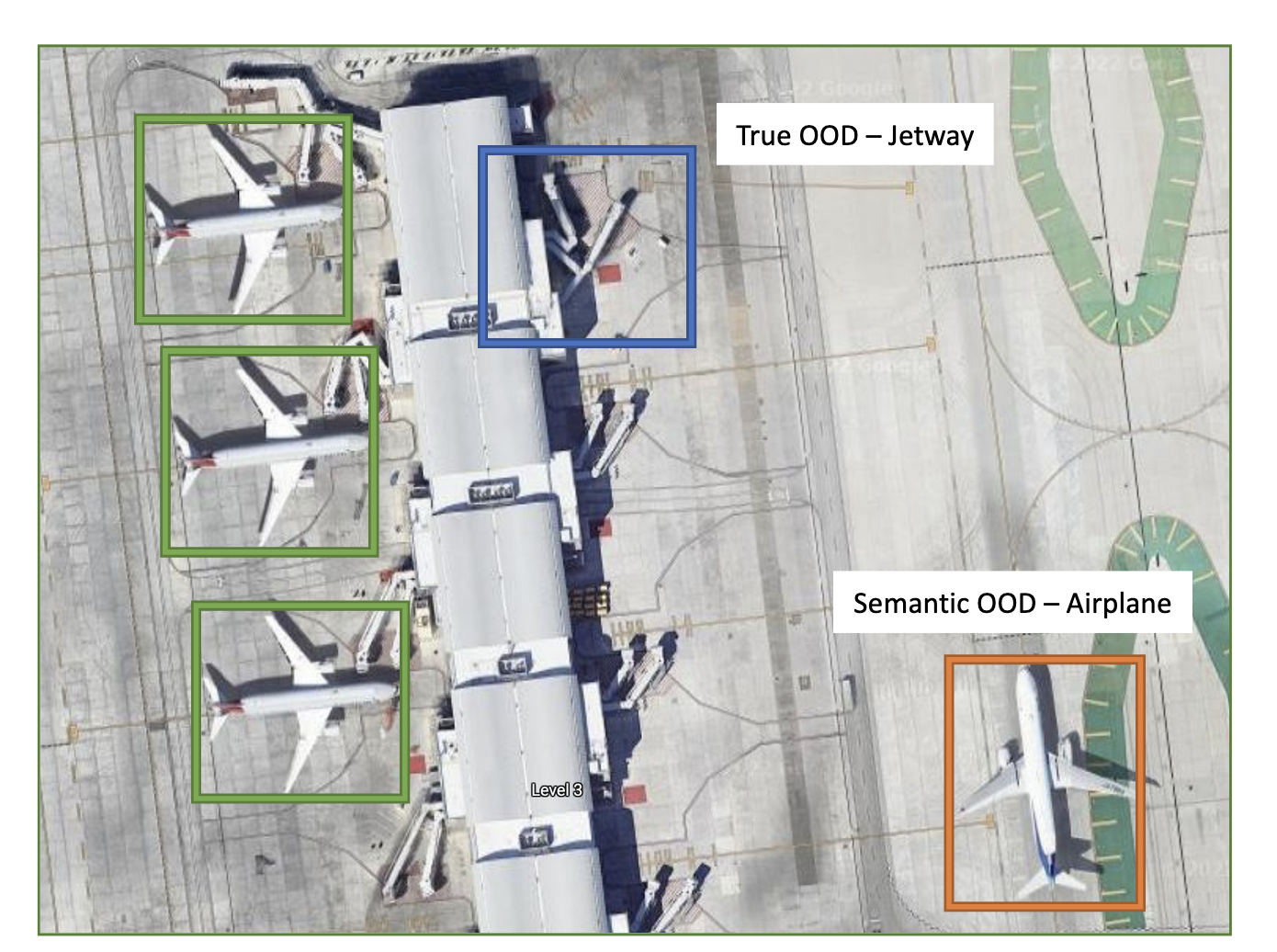}
\end{tabular}
\end{center}
\caption 
{Our hypothetical object-detector and classifier model can produce two distinct types of OOD errors. An error by the object detector in picking up a section of jetway in blue is a `True' OOD error, while an out of distribution plane for the classifier in orange is a `Semantic' OOD error. Image from \cite{googlemaps}}
\label{fig:LAXanno}
\end{figure}

Now let us propose a change of frame: consider the upgraded task of a paired object detector and classifier for satellite imagery of airplanes on a runway, where the detector selects objects of interest, and passes them to the classifier for identification. The object detector only feeds the classifier what it identifies as an airplane, and any other objects in the image are ignored. In this application, our concern is not with fine- versus coarse-grained OOD but with semantically meaningful OOD versus `True' OOD. As shown in \ref{fig:LAXanno}, if the object detector were to mistakenly identify a section of jetway as an airplane and that error propagated into being labeled as a specific model of airplane, that is a meaningfully different type of OOD error than one where an OOD airplane is correctly detected but misidentified as a known airplane.

Now consider a hypothetical over-exuberant model, measured using the hierarchical AUROC/FPR metrics outlined above, that consistently rates both fine- and coarse-grained OOD with low confidence scores indiscriminately. Such a model would treat semantically meaningful OOD airplanes the same as the True OOD jetway, despite scoring highly on the above hierarchical metric. Similar to how accuracy can hide low precision or recall on highly unbalanced data sets, hierarchical AUROC/FPR on its own can hide a model mistaking important OOD samples (OOD airplanes) from unimportant ones (sections of jetway). Semantic vs. True metrics detect this and can therefore contextualize our hierarchical metrics. 

The Semantic vs True OOD conceptualization reveals an important distinction between the fine-grained task and much of the existing research into coarse-grained OOD detection. Existing methods are evaluated on a binary task: discerning in-distribution samples from out-of-distribution samples. This approach is overly simplistic in a similar way to the problem it professes to solve. Applications of binary-task OOD methods will miss the distinction between Semantic and True OOD on a structural level just as applications of closed-world models in the open-world will miss the potentially infinite set of classes on which they have not been trained.

\begin{figure}[h]
\begin{center}
\begin{tabular}{c}
\includegraphics[width=0.8\linewidth]{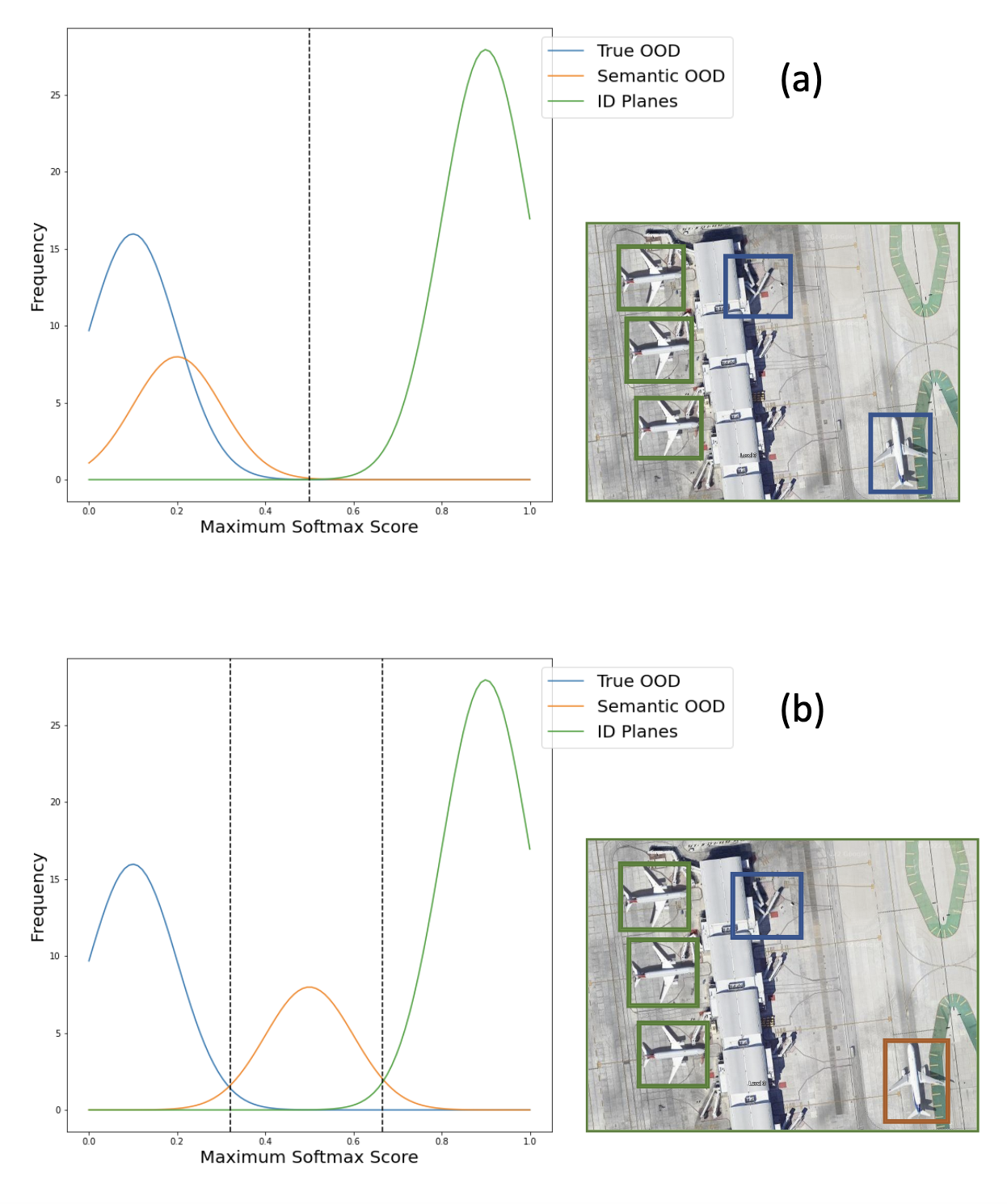}
\end{tabular}
\end{center}
\caption{ (a) shows the the frequency of the maximum softmax score in an over-exuberant model with poor STAUROC but strong Hierarchical AUROC, while (b) shows the the frequency of the maximum softmax score between the three OOD data sets for an idealized model with near-perfect STAUROC and Hierarchical AUROC scores}
\label{fig:idealternarydensity}
\end{figure} 

From here, we can re-imagine this as a problem of ternary classification. Now the goal of an Maximum Softmax Probability (MSP) detector is to set two thresholds. Predictions with confidence above the higher threshold are classed as ID, between the two thresholds as Semantic OOD, and below both thresholds as True OOD. Figure \ref{fig:idealternarydensity} shows how an idealized model might separate the Max Softmax scores of the three sets. With the problem restated like this, we can imagine the hierarchical AUROC/FPR measuring the separation between the ID and Semantic OOD sets. To fully evaluate the performance, we propose measuring the AUROC/FPR between the Semantic and True sets (STAUROC/STFPR), i.e. the performance of the lower of the two thresholds, shown on the left in Fig. \ref{fig:idealternarydensity}. The hypothetical model above that consistently rates fine- and coarse-grained OOD with low confidence scores will have a high hierarchical AUROC but a low STAUROC.

\subsection{TernaryMixOE}

The existing MixOE-Linear and MixOE-Cut methods outlined in \cite{mixup} propose the use of \textit{mixing operations}, such that, for $X_{in}=(x_{in}, y_{in}) \sim \mathcal{D}_{in}$ in the ID training set and $X_{out}=(x_{out}, y_{out}) \sim \mathcal{D}_{out}$ in the OOD training set and a given a mixing coefficient $\lambda\in[0,1]$ randomly generated on a per-batch basis, we have

\begin{equation*}
    x_{in/out} = \mix(x_{in}, x_{out}, \lambda)
\end{equation*}

and the correspondingly blended targets drawn from a Uniform Distribution, $\mathcal{U}$

\begin{equation*}
    y_{in/out} = \lambda y_{in} + (1-\lambda)\mathcal{U}
\end{equation*}

creating a `virtual out' data set $D_{in/out}$, which allows for improved fine-grained performance by interpolating ID and OOD samples when applied with the following training objective, for a model $f$, cross-entropy $\mathcal{L}$ and a weighting hyperparameter $\beta$



\begin{multline*}
    \mathbb{E}_{X_{in} \sim \mathcal{D}_{in}}[ \mathcal{L}(f(x), y)] + \\
    \beta \mathbb{E}_{X_{in/out}\sim \mathcal{D}_{in/out}}[ \mathcal{L} (f(x_{in/out}), y_{in/out})]
\end{multline*}

We propose, as an extension of this method, the addition of a third loss term that blends two ID samples. Just as MixOE-Linear and MixOE-Cut create a `virtual out' training set, our method, TernaryMixOE, creates a  `virtual in' in a similar manner. Given two in distribution samples $X=(x, y), X'=(x', y') \sim \mathcal{D}_{in}$ and a mixing coefficient $\lambda\in[0,1]$, we have

\begin{equation*}
\begin{aligned}
    x_{in/in} &= \mix(x, x', \lambda) \\
    y_{in/in} &= \lambda y + (1-\lambda)y'
\end{aligned}
\end{equation*}

The intuition behind this addition is that, much like the existing mix methods encourage interpolation between the in and out training sets, fine-grained OOD examples are more likely to exist visually between known ID classes. Therefore, training in this space between in-distribution examples should increase performance at the finest level. 

Likewise, we expand the training objective of the mix methods, with the addition of a second weighting hyperparameter, $\gamma$.

\begin{equation*}
\begin{split}
\mathbb{E}_{X_{in}\sim \mathcal{D}_{in}}&[\mathcal{L}(f_{in}(x), y_{in})] \\
    &+ \beta \mathbb{E}_{X_{in/out}\sim \mathcal{D}_{in/out}}[\mathcal{L}(f_{in/out}(x), y_{in/out})]\\
    &+ \gamma \mathbb{E}_{X_{in/in}\sim \mathcal{D}_{in/in}}[\mathcal{L}(f_{in/in}(x), y_{in/in})]
\end{split}    
\end{equation*}

\section{Experiments}
\label{sec:experiments}

\subsection{Setup}

We set baselines for five state-of-the-art OOD methods: Baseline \cite{Vanilla}, Outlier Exposure (with \cite{OE} and without \cite{ODIN} temperature scaling), Energy \cite{Energy}, and MixOE-Cut \cite{mixup}, in addition to our TernaryMixOE method. For each experiment we used ImageNet pre-trained ResNet-50 models \cite{resnet50} followed by a single linear layer. Each model was trained with a batch size of 20 for 100 epochs. For the training outlier set, we used ImageNet 2010 of 1.26 million natural images taken from the internet \cite{ILSVRC15}. Training was performed on a server of 4 Tesla V100 GPUs.

The image transformation process is similar to \cite{mixup}. Before training, images are resized to $512\times512$ before being randomly cropped to $448\times448$, randomly flipped along their horizontal axis, and given a random color jitter. Finally, the images are normalized according to the ImageNet mean and standard deviation \cite{ILSVRC15}.

We performed a hyperparameter search on each method, and with the exception of energy we saw no changes needed from those recommended in their respective papers. For Energy, we tested the Margin-in parameter and found a steep drop off in Semantic vs. True OOD performance below 25, for this reason we adjusted the Margin-in parameter to 25.

To perform the Semantic versus True AUROC/FPR evaluation outlined in Sec. \ref{sec:stauroc}, we created a combination dataset of Stanford Dogs \cite{dogs}, Places365 \cite{places}, and Stanford Cars \cite{cars} for coarse grained evaluation relative to our fine-grained hold out sets.

\subsection{Evaluation}

Our evaluation procedure involves measuring the class-accuracy on the in-distribution test set and the overall AUROC/FPR between ID test set and the fine-grained hold out set, in addition to our two metrics outlined in Sec. \ref{sec:evalmetrics}. Table \ref{tab:idacc} shows the in-distribution accuracy for the five training methods. Our first observation is the increase in performance for mix operation methods MixOE-Cut and TernaryMixOE compared to the other methods, particularly Baseline + MSP with no outlier exposure at all. This result is notable, as pure outlier-exposure methods like OE and Energy see a degradation in ID accuracy relative to the baseline method without outlier exposure. This may hint at the ability of mix methods to improve performance in closed-world applications, where OOD detection is not needed.

\begin{table}[ht]
\centering
\begin{adjustbox}{width=\linewidth}
\begin{tabular}{|l|l|c|}
\hline\rule[-1ex]{0pt}{6ex} 
Data set & \begin{tabular}[c]{@{}l@{}}Method \\ (Loss / Detector)\end{tabular} & ID Accuracy    \\
\hline\rule[-1ex]{0pt}{4ex} 
\multirow{6}{*}{\begin{tabular}[c]{@{}l@{}}Average of \\ FGVC-A Splits / \\ ImageNet\end{tabular}}
        & TernaryMixOE / MSP                                                    & 93.22             \\
        & MixOE-Cut / MSP                                                       & \textbf{93.23}    \\
        & OE / MSP                                                              & 91.06             \\
        & Baseline / MSP                                                        & 92.05             \\
        & Energy / Energy                                                       & 90.71             \\
\hline\rule[-1ex]{0pt}{4ex} 
\multirow{6}{*}{\begin{tabular}[c]{@{}l@{}}Ships Military Split / \\ ImageNet\end{tabular}} \
        & TernaryMix OE / MSP                                                 & \textbf{95.55} \\
        & MixOE-Cut / MSP                                                     & 95.15          \\
        & OE / MSP                                                            & 94.12          \\
        & Baseline / MSP                                                      & 93.81          \\
        & Energy / Energy                                                     & 94.16          \\
\hline
\end{tabular}
\end{adjustbox}
\vspace{5pt}
\caption{In-distribution Accuracy of each of the five training methodologies on an average of the FGVC-Aircraft Splits and on the ShipsRSImageNet Military Split}
\label{tab:idacc}
\end{table}

As for our Hierarchical AUROC metric, shown in Table \ref{tab:hierauroc}, we come away with three key observations. 

\begin{enumerate}
    \item Performance degrades as the OOD test set moves deeper into the hierarchy. On both data sets and for all models, the performance at the finest-grained hold-out set of significantly worse than the most coarse-grained hold-out set. This verifies our intuition about the hierarchical metrics outlined in Sec. \ref{sec:evalmetrics}.
    \item For Ships, Mix methods like TernaryMixOE and MixOE-Cut under-perform standard Outlier exposure and Outlier exposure with temperature scaling in all metrics. We hypothesize two possible reasons for this behavior: a) the vast different in resolution between the most and least detailed ships in the in-distribution set makes the task unsuitable for Mix methods, and b) the change in point-of-view between the in-distribution set (overhead satellite imagery) and the training OOD set (mostly human POV) degrades performance of the Mix methods but not regular outlier exposure. Further experimentation is needed to confirm this, regardless TernaryMixOE outperforms MixOE-Cut overall and in both levels of the hierarchy.
    \item Most notably, our method, TernaryMixOE performs worse than MixOE-cut at Level 1 AUROC/FPR and on the overall AUROC/FPR for FGVC-A, but performs better specifically at the finest-grained classification against all existing baselines. This proves it is possible for a model to isolate a performance improvement to finer-grained OOD detection when compared to a baseline. Future research can use hierarchical metrics to investigate whether this trend can be seen in other models.
\end{enumerate}

\begin{table*}[ht]
\centering
\begin{adjustbox}{width=1\textwidth}
\begin{tabular}{|l|l|c|c|c|c|}
\hline\rule[-1ex]{0pt}{6ex} 
Data set &
  \begin{tabular}[c]{@{}l@{}}Method \\ (Loss / Detector)\end{tabular} &
  \begin{tabular}[c]{@{}l@{}}L1 AUROC/\\ FPR\end{tabular} &
  \begin{tabular}[c]{@{}l@{}}L2 AUROC/\\ FPR\end{tabular} &
  \begin{tabular}[c]{@{}l@{}}L3 AUROC/\\ FPR\end{tabular} &
  \begin{tabular}[c]{@{}l@{}}All AUROC/\\ FPR\end{tabular} \\
\hline\rule[-1ex]{0pt}{4ex} 
\multirow{7}{*}{\begin{tabular}[c]{@{}l@{}}Average of \\ FGVC-A Splits / \\ ImageNet\end{tabular}} 
 & TernaryMixOE / MSP               & 93.64 / 21.11 & 88.78 / 41.49 & \textbf{80.81} / \textbf{49.93}   & 88.86 / 43.27 \\
 & MixOE-Cut / MSP                  & \textbf{94.93} / \textbf{13.77} & \textbf{90.51} / \textbf{33.19} & 79.31 / 59.38 & \textbf{89.63} / \textbf{40.88} \\
 & OE / MSP                         & 88.37 / 33.03 & 87.46 / 39.08 & 72.63 / 75.33 & 84.19 / 54.64 \\
 & OE / MSP (t=1000)                & 89.09 / 36.62 & 85.70 / 49.96 & 62.93 / 89.16 & 81.45 / 73.20 \\
 & Baseline / MSP                   & 89.14 / 32.56 & 88.43 / 38.88 & 75.39 / 68.28 & 85.62 / 50.94 \\
 & Energy / Energy                  & 88.39 / 42.08 & 83.18 / 62.95 & 68.24 / 84.68 & 81.62 / 69.72 \\
\hline\rule[-1ex]{0pt}{4ex} 
\multirow{7}{*}{\begin{tabular}[c]{@{}l@{}}Ships Military Split / \\ ImageNet\end{tabular}} 
 & TernaryMixOE / MSP               & 95.52 / 17.67                     & 82.98 / 72.87          & & 93.77 / 28.27                   \\
 & MixOE-Cut / MSP                  & 93.99 / 28.85                     & 79.51 / 83.11          & & 91.96 / 50.00                   \\
 & OE / MSP                         & 96.95 / 13.98                     & 85.48 / \textbf{58.31} & & 95.35 / 21.75                   \\
 & OE / MSP (t=1000)                & \textbf{97.86} / \textbf{11.15}   & \textbf{85.81} / 61.71 & & \textbf{96.17} / \textbf{22.04} \\
 & Baseline / MSP                   & 89.54 / 41.03                     & 76.87 / 88.41          & & 87.77 / 53.11                   \\
 & Energy / Energy                  & 85.20 / 52.02                     & 70.66 / 87.86          & & 83.17 / 62.10                   \\
\hline

\end{tabular}
\end{adjustbox}
\vspace{5pt}
\caption{Hierarchical AUROC/FPR for the six training method/detector combinations, in addition to the overall AUROC/FPR. For FGVC-A, Level 1 refers examples held out at the manufacturer hierarchy level, Level 2 at the model level, and Level 3 at the variant level. For ShipsRSImageNet, Level 1 refers to Military OOD, and Level 2 refers to the held out `other civilian' catch-all class.}
\label{tab:hierauroc}
\end{table*}

\begin{table*}[ht]
\centering
\begin{adjustbox}{width=0.66\textwidth}
\begin{tabular}{|l|l|c|}
\hline\rule[-1ex]{0pt}{6ex}
Data set &
  Method (Loss / Detector) &
  \begin{tabular}[c]{@{}l@{}}Semantic vs. True \\ AUROC/FPR\end{tabular} \\
\hline\rule[-1ex]{0pt}{4ex}
\multirow{7}{*}{\begin{tabular}[c]{@{}l@{}}Average of\\ FGVC-A / \\ ImageNet\end{tabular}}  
 & TernaryMixOE / MSP       & 92.93 / 34.81                     \\
 & MixOE-Cut /  MSP         & 99.27 / 1.64                      \\
 & OE / MSP                 & \textbf{100.00} / \textbf{0.00}   \\
 & OE / MSP (t=1000)        & \textbf{100.00} / \textbf{0.00}   \\
 & Baseline / MSP           & 81.96 / 78.23                     \\
 & Energy / Energy          & 67.61 / 91.41                     \\
\hline\rule[-1ex]{0pt}{4ex}
\multirow{7}{*}{\begin{tabular}[c]{@{}l@{}}Ships Military Split / \\ ImageNet\end{tabular}}
 & TernaryMixOE / MSP       & 96.58 / 14.59                     \\
 & MixOE-Cut / MSP          & 96.06 / 20.97                     \\
 & OE / MSP                 & 98.59 / 6.02                      \\
 & OE / MSP (t=1000)        & \textbf{98.61} / \textbf{5.98}    \\
 & Baseline / MSP           & 65.29 / 79.06                     \\
 & Energy / Energy          & 83.75 / 59.19                     \\
\hline
\end{tabular}
\end{adjustbox}
\vspace{5pt}
\caption{Semantic vs True AUROC/FPR for the six training method/detector combinations}
\label{tab:fvc}
\end{table*}

For our Semantic vs. True AUROC/FPR metrics, shown in Table \ref{tab:fvc}, the most immediate result is the perfect separation of outlier exposure methods in FGVC-A and the near perfect performance in ShipsRSImageNet. Given their weak performance on the hierarchical metric, we can say OE models have the opposite problem of our hypothetical over-exuberant model outlined in Sec. \ref{sec:evalmetrics}, where in our results the lower threshold between the semantic and true OOD data sets under-performs the upper threshold. Secondly with respect to TernaryMixOE and MixOE-Cut, we observe a correlation in the degradation in performance in STAUROC and the under-performance of TernaryMixOE in the Hierarchical AUROC. In FGVC-A, TernaryMixOE under-performs except at the finest grained level while degrading performance six points in the STAUROC metric, but no such loss is observed for ShipsRSImageNet where TernaryMixOE performs better at all levels of the Hierarchical metric. This may hint that the performance of the two metrics is linked, and that as fine-grained classification performance improves, models will naturally tend toward a three-way separation of the data sets. As more methods are developed and evaluated using these metrics, we will be able to observe whether this holds in future research.

\section{Conclusion}

In this paper, we re-conceptualized Fine-grained OOD classification into separate in- and out-of-distribution granularity systems. Additionally, we introduced two new metrics to evaluate the fine-grained performance of OOD models and described an extension of MixOE that can isolate a performance improvement to the finest-grained classification level. Future research can expand on our results by formalizing or quantifying our understanding of granularity, and identifying  new methods to improve fine-grained performance.

\bibliography{main.bib} 
\bibliographystyle{bib_style.bst}

\end{document}